\documentclass[letterpaper]{article} 
\usepackage{aaai2027}  
\usepackage[hyphens]{url}  
\usepackage{graphicx} 
\usepackage{natbib}  
\usepackage{caption} 
\usepackage{algorithm}
\usepackage{algorithmic}
\usepackage{makecell}
\usepackage{amssymb}
\usepackage{amsmath}
\usepackage{xcolor}
\usepackage{tabularx}
\usepackage{makecell}
\usepackage{pdfpages}

\definecolor{gaincolor}{rgb}{0.1, 0.5, 0.2} 

\definecolor{gaincolor}{HTML}{228B22} 

\usepackage{newfloat}
\usepackage{listings}
\DeclareCaptionStyle{ruled}{labelfont=normalfont,labelsep=colon,strut=off} 
\floatstyle{ruled}
\newfloat{listing}{tb}{lst}{}
\floatname{listing}{Listing}

\usepackage{booktabs}

\title{Beyond Global Realism: Virtual Try-On Evaluation and Optimization with Dimension-wise Garment Fidelity Assessment}
\author{
    Kaidong Zhang\textsuperscript{\rm 1}\equalcontrib,
    Yukang Ding\textsuperscript{\rm 1}\equalcontrib,
    Xiaoyu Liu\textsuperscript{\rm 2},
    Ying Chen\textsuperscript{\rm 1}\corresponding
}
\affiliations{
    \textsuperscript{\rm 1}Taobao \& Tmall Group of Alibaba, 
    \textsuperscript{\rm 2} Harbin Institute of Technology \\
    zhangkaidong.zkd@taobao.com
}

\nocopyright

\begin{document}

\maketitle

\begin{abstract}
Virtual try-on (VTON) requires not only realistic generation but also faithful preservation of garment characteristics. 
However, existing evaluation metrics such as PSNR, SSIM, KID and FID struggle to measure the consistency between the generated and reference garments, particularly in capturing the multi-dimensional characteristics of garment fidelity.
To address this, we propose \textbf{DAT}: a \textbf{D}imension-wise \textbf{A}ssessment framework for virtual \textbf{T}ry-on, which decomposes garment consistency into seven interpretable dimensions: silhouette, color, neckline and sleeve shape, major decoration and structure, material texture, fine-detail fidelity, and logo preservation, each formulated as a discrete attribute-level prediction task.
To train this specialized assessment model, we adopt a two-stage learning paradigm comprising large-scale weak supervision on 50K samples, followed by refinement on 10K higher-quality annotations obtained via multi-model voting. 
Furthermore, we employ weighted cross-entropy loss to mitigate the severe label imbalance inherent across evaluation dimensions.
Beyond its role as an evaluation framework, the assessment model can be integrated into reinforcement learning optimization of Qwen-Image-Edit for VTON, where dimension-wise rewards are adaptively aggregated to emphasize under-optimized aspects during training.
Experimental results show that our method (8B parameters) achieves state-of-the-art performance in terms of balanced accuracy, SROCC, and PLCC, outperforming strong proprietary models such as Gemini-3.1, Qwen3.7-plus, and GPT-5.5, while also serving as an effective optimization signal for reward-guided VTON generation.
\end{abstract}

\section{Introduction}

Virtual try-on (VTON) aims to transfer a target garment onto a person image while preserving both visual realism and garment fidelity, and has broad applications in e-commerce and digital fashion~\cite{cordier20012d,han2018viton,wang2018toward,hsieh2019fashionon}. Although recent image editing models~\cite{firered2026rededit,wu2025qwen,nanobanana_pro} have substantially improved the realism of generated results, realistic synthesis alone is insufficient in practice. A useful try-on image must also faithfully preserve the source garment, including its silhouette, color, local structure, texture, and brand-related details. Even minor deviations in these attributes may significantly reduce the commercial reliability of the generated result.
However, garment consistency assessment in VTON remains underexplored. Existing metrics such as PSNR~\cite{gonzalez2002digital}, SSIM~\cite{ssim2004}, FID~\cite{heusel2017gans}, KID~\cite{binkowski2018demystifying}, CLIP-I~\cite{radford2021learning}, DiffSim~\cite{song2025diffsim} and LPIPS~\cite{zhang2018perceptual} are designed for global image quality evaluation and thus fail to capture fine-grained garment attributes. In practice, garment fidelity is inherently multi-dimensional: a try-on result may faithfully reproduce the overall color while distorting the neckline, or preserve the silhouette while losing logos and surface details. Collapsing such diverse attributes into a single scalar score obscures attribute-level discrepancies, limiting interpretability and reducing utility for model diagnosis, result ranking, and downstream optimization.

To address this limitation, we propose \textbf{DAT}: a \textbf{D}imension-wise \textbf{A}ssessment framework for virtual \textbf{T}ry-on. Instead of predicting a single overall score, DAT evaluates garment consistency along seven dimensions: \textit{silhouette}, \textit{color}, \textit{neckline and sleeve shape}, \textit{major decoration and structure}, \textit{material texture}, \textit{fine-detail fidelity}, and \textit{logo preservation}. Each dimension is modeled as a discrete attribute-level prediction task. For dimensions whose corresponding attribute may be absent, such as fine-detail fidelity and logo preservation, we introduce an additional not-applicable (NA) category. This formulation enables more informative and interpretable assessment than black-box scalar modeling. Such dimension design is motivated by practical business requirements for garment fidelity assessment, covering both global appearance and identity-critical local details.

Training such assessment model is challenging due to annotation noise and severe score imbalance across dimensions. To tackle annotation noise, we adopt a two-stage training framework. Based on Qwen3-vl-8b model~\cite{bai2025qwen3}, we first train DAT from 50K weakly labeled samples annotated by Gemini-3.1~\cite{gemini_31_pro}, and then refine the model on 10K higher-quality samples constructed through multi-model voting among Gemini-3.1~\cite{gemini_31_pro}, Qwen3.7-plus~\cite{qwen37_plus} and GPT-5.5~\cite{gpt55}. To mitigate the severe imbalance in dimension-wise score labels, we employ weighted cross-entropy loss to increase the contribution of underrepresented score categories during training. Moreover, the dimension-wise assessment from DAT can serve as a powerful reward signal to improve the VTON models. We utilize DAT as the reward model and adopt Flow-GRPO~\cite{liu2026flow} method to perform reinforcement learning (RL) optimization of Qwen-Image-Edit~\cite{wu2025qwen}. Different from simple average of the rewards from multiple dimensions , we introduce an adaptive aggregation strategy that dynamically increases the weights of under-optimized dimensions during training, encouraging more balanced improvement across garment attributes.

Extensive experiments show that DAT outperforms strong proprietary judges, including Gemini-3.1~\cite{gemini_31_pro}, Qwen3.7-plus~\cite{qwen37_plus}, and GPT-5.5~\cite{gpt55}, across VTON results from multiple sources~\cite{wu2025qwen,firered2026rededit,nanobanana_pro,nanobanana2,gpt_image2} on balanced accuracy, SROCC, and PLCC. Notably, DAT has only 8B parameters, making it substantially more compute- and resource-efficient than these proprietary large models. 
Moreover, reward-guided RL optimization based on DAT further improves reward-aligned garment preservation over the original Qwen-Image-Edit~\cite{wu2025qwen} model and yields qualitatively more faithful try-on results.

Our contributions are three-fold:
\begin{itemize}
    \item We propose DAT, a multi-dimensional garment assessment framework for VTON that decomposes garment fidelity into seven interpretable dimensions, enabling more informative and interpretable evaluation than conventional scalar metrics.
    
    \item We design a two-stage, imbalance-aware training framework for learning the assessment model under noisy supervision, leveraging large-scale weak labels, multi-model-voted refinement data, and weighted cross-entropy to handle annotation noise and severe score imbalance.
    
    \item We show that the learned 8B-parameter model is effective for both assessment and optimization: it outperforms strong proprietary judges and can be integrated into reinforcement learning framework as a multi-dimensional reward to improve garment-faithful VTON generation.
\end{itemize}

\section{Related Work}

\paragraph{Virtual Try-On and Its Evaluation Metrics}
VTON aims to generate an image of a person wearing a target garment while preserving both human realism and garment identity.~\cite{choi2021viton,Cui_2021_ICCV,chen2021fashionmirror,ge2021disentangled}. Early methods mainly rely on pose estimation, human parsing, geometric warping, and garment-to-body alignment to transfer target clothing onto a person image~\cite{Li_2023_ICCV,xie2023gpvton}. More recent approaches leverage diffusion models and large image editing models to achieve more realistic and controllable synthesis~\cite{firered2026rededit,wu2025qwen,Yang_2024_CVPR}. Despite this progress, most existing methods primarily focus on visual realism, pose compatibility, and overall generation quality~\cite{ning2024picture,gou2023taming}. Current methods mainly utilize evaluation metrics that focus on pixel similarity, perceptual distance, or global vision-language alignment, such as SSIM~\cite{1284395}, LPIPS~\cite{zhang2018perceptual,ghildyal2022stlpips} and CLIP-I~\cite{hessel2021clipscore}. Such metrics excel at global consistency but often overlook localized discrepancies. In contrast, the problem of faithfully preserving garment identity, especially at a fine-grained attribute level, has received much less attention. Our work complements prior virtual try-on research by focusing on garment consistency assessment rather than generation alone.


\paragraph{Reward Modeling and Preference Learning}
Reward modeling and preference learning have become important tools for aligning generative models with human preferences~\cite{christiano2017deep,ouyang2022training,parksurf}. Existing approaches typically learn a reward function from pairwise comparisons, ranking signals, or model-generated annotations, and have been widely explored in language generation, image generation, and multimodal alignment~\cite{kim2023aligning,wu2025rewarddance,zhang2025r1}. However, most reward models produce a single scalar score for each sample~\cite{luo2025editscore,lai2026posterreward}, which is suitable for coarse preference judgment but limited for tasks requiring attribute-level assessment. Some prior methods also explore attribute-level analysis or local detail matching in fine-grained vision tasks~\cite{10262331,yang2026finegrained,cao2025artimusefinegrainedimageaesthetics,sheng2026fgresq,zhang2023liqe}, yet they are not designed for reward modeling in virtual try-on, nor do they provide interpretable multi-dimensional supervision for downstream optimization. In virtual try-on, garment fidelity is inherently multi-dimensional, and a scalar evaluation metric cannot explain which aspects of the garment are preserved or distorted. Different from prior methods, our method explicitly conducts multi-dimensional evaluation and predicts fine-grained garment-related rewards for virtual try-on.

\section{Method}

\subsection{Problem Formulation}

\begin{figure*}[!htbp]
\centering
\includegraphics[width=0.99\textwidth]{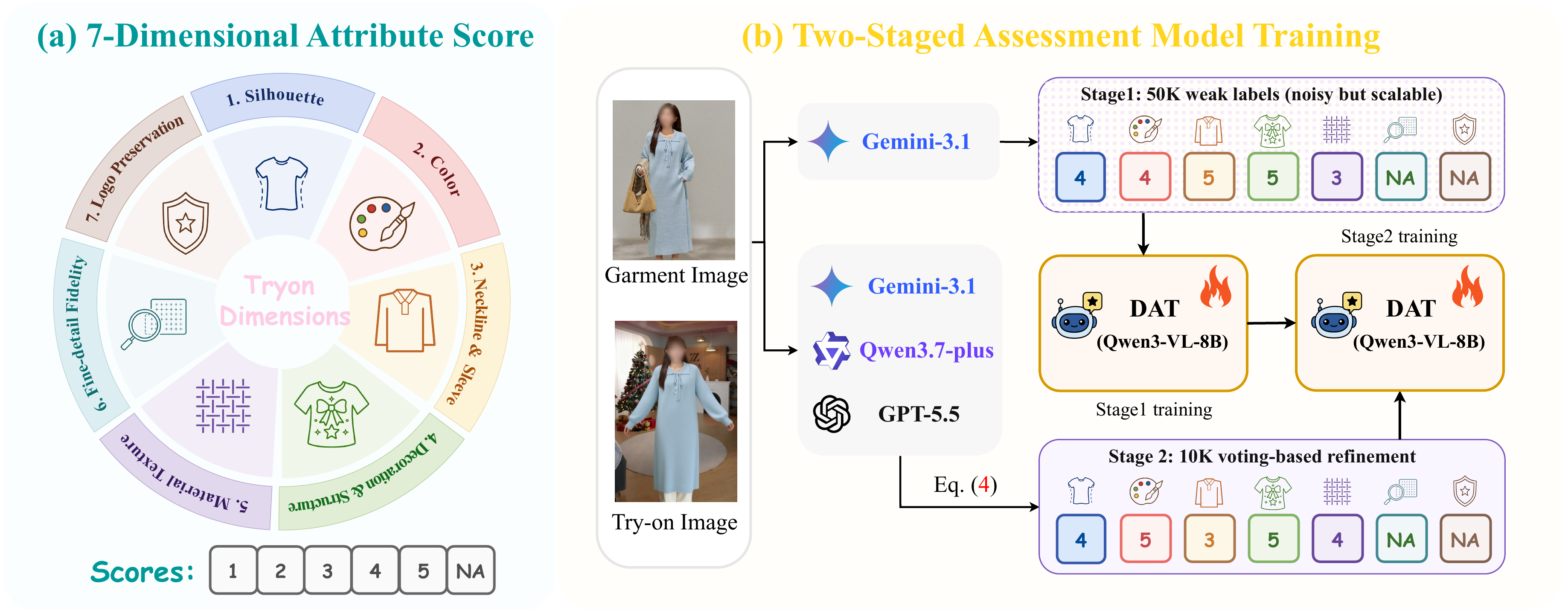} 
\caption{The illustration of the 7-dimensional attribute score (a) and the training pipeline (b) in DAT.}
\label{fig:method}
\vspace{-0.5cm}
\end{figure*}

Given a garment image $I_g$, a person image $I_p$, and a virtual try-on result $I_t$, our goal is to learn an assessment model that evaluates how faithfully the generated image preserves the garment characteristics specified by $I_g$. Unlike conventional models that predict a single holistic scalar, we formulate VTON assessment modeling as a fine-grained multi-dimensional assessment problem.

Built upon Qwen3-VL-8B~\cite{bai2025qwen3}, our model takes the garment image $I_g$ and the try-on result $I_t$ as inputs and predicts attribute-level consistency scores over seven predefined dimensions: silhouette, color, neckline and sleeve shape, major decoration and structure, material texture, fine-detail fidelity, and logo preservation. The details of these dimensions are presented in the supplementary material. Formally, each sample is associated with a label vector
\begin{equation}
\mathbf{y} = [y_1, \dots, y_7],
\end{equation}
where
\begin{equation}
y_d \in \{1,2,3,4,5,\texttt{NA}\}, \quad d=1,\dots,7.
\end{equation}
We illustrate the proposed 7-dimensional attribute score in Fig.~\ref{fig:method} (a). Larger scores indicate better garment consistency, while $\texttt{NA}$ denotes that the corresponding attribute is not applicable or cannot be reliably judged for the current sample.

The assessment modeling task is then to learn a function
\begin{equation}
f_\theta(I_g, I_t) = \{\mathbf{p}_d\}_{d=1}^7,
\end{equation}
where each $\mathbf{p}_d \in \mathbb{R}$ denotes the predicted probability distribution over the label vector for attribute dimension $d$.

\subsection{Assessment Model Training}

A key challenge in training multi-dimensional assessment models for virtual try-on lies in the tension between annotation scale and annotation fidelity. Since each sample requires judgments on seven garment-consistency attributes, obtaining large-scale high-quality supervision is prohibitively expensive. Meanwhile, relying solely on a single automatic annotator is cost-effective but introduces non-negligible noise and model-specific bias, particularly on subtle dimensions such as texture, fine details, and logo preservation. In addition, the multi-dimensional score distribution is highly imbalanced, making naive empirical risk minimization prone to overfitting dominant score levels. To address these issues, we adopt a staged supervision strategy together with imbalance-aware optimization. We show the training pipeline of the assessment model in Fig.~\ref{fig:method} (b).

\paragraph{Stage 1: Large-scale weak supervision.}
In the first stage, we use Gemini-3.1~\cite{gemini_31_pro} to annotate large-scale triplets $(I_g, I_p, I_t)$ with seven dimension-wise scores. The purpose of this stage is to construct a coarse but usable supervision source at manageable cost, so the model can learn broad attribute-level correspondences between the reference garment and the generated try-on image across diverse garment categories and generation qualities. Although single-model annotations are inevitably noisy, they provide sufficient scale and coverage for learning a strong initialization.

\paragraph{Stage 2: Consensus-based low-noise refinement.}
To obtain more reliable supervision, we further build a higher-quality refinement set from a smaller set of samples. For each sample and each attribute dimension $d$, we utilize Gemini-3.1, Qwen3.7-plus, and GPT-5.5 independently to produce three ordinal scores, denoted by $s_d^{(1)}, s_d^{(2)}, s_d^{(3)}$. We aggregate them into the final label by
\begin{equation}
\tilde{y}_d = \operatorname{median}\!\left(s_d^{(1)}, s_d^{(2)}, s_d^{(3)}\right).
\end{equation}
For three annotators, this is equivalent to the rule \emph{``2:1 majority wins, otherwise take the median''}. Such median-based aggregation is particularly suitable for ordinal labels, as it suppresses outlier predictions while preserving the order structure of the score space. We then continue training the stage-1 model on this voting-based refinement set, thereby improving calibration and reducing the impact of annotator-specific noise on difficult attributes.

\begin{figure*}[!htbp]
\centering
\includegraphics[width=0.99\textwidth]{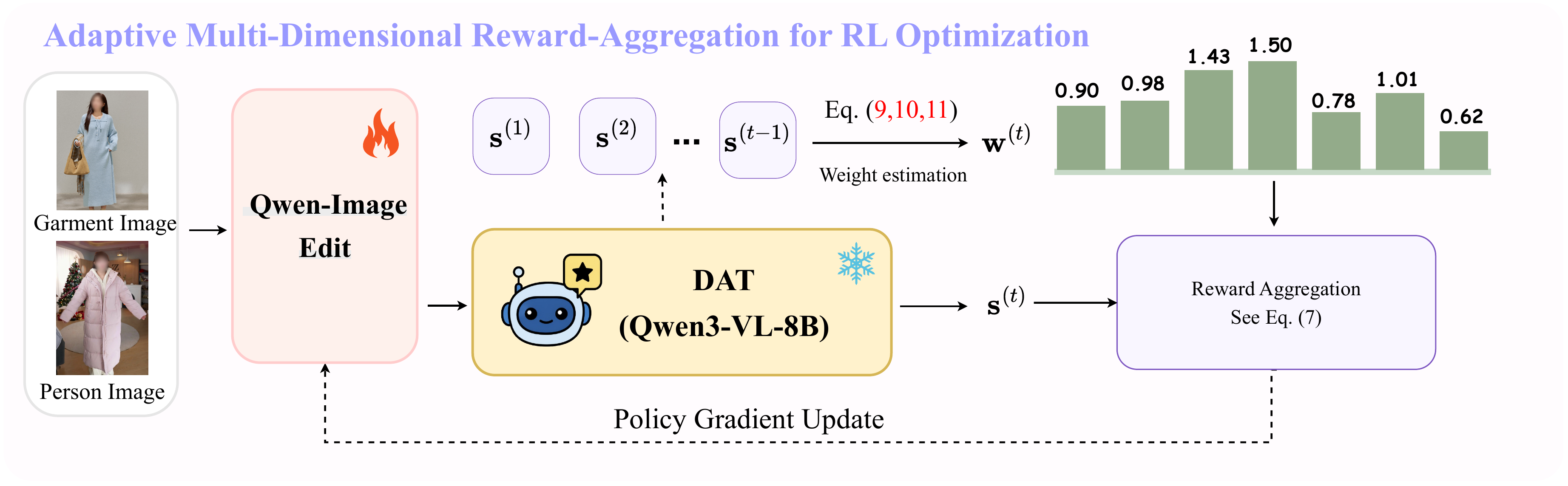} 
\vspace{-0.3cm}
\caption{The framework of the adaptive multi-dimensional reward-aggregation RL optimization. We utilize DAT as the reward model to provide the reward signal.}
\label{fig:method_rl}
\vspace{-0.5cm}
\end{figure*}

\paragraph{Imbalance-aware optimization.}
A further difficulty comes from the highly skewed score distribution across both attribute dimensions and score levels. In principle, one may attempt to rebalance the dataset by resampling underrepresented scores. However, this is unsuitable in our setting: balancing the score distribution of one attribute dimension would generally distort the distributions of other dimensions, since each sample is simultaneously annotated along all seven attributes. We therefore handle imbalance at the loss level rather than the data level, by increasing the training contribution of rare score categories.

Let $q_{d,c}$ denote the empirical probability of score level $c \in \{1,\dots,5,\texttt{NA}\}$ under attribute dimension $d$ in the training set. We define the corresponding class weight as
\begin{equation}
\alpha_{d,c} = \operatorname{clip}\!\left(\frac{1}{q_{d,c}}, \alpha_{\min}, \alpha_{\max}\right),
\end{equation}
where $\operatorname{clip}(x,\alpha_{\min},\alpha_{\max})$ truncates $x$ into the interval $[\alpha_{\min}, \alpha_{\max}]$. We empirically set $\alpha_{min}$ and $\alpha_{max}$ to 1 and 3, respectively. This inverse-frequency weighting amplifies the gradient contribution of underrepresented scores, while clipping prevents extremely rare categories from introducing unstable or disproportionately large updates.

Based on these weights, we optimize the assessment model using a validity-aware weighted cross-entropy loss:
\begin{equation}
\mathcal{L}_{\text{RM}} = \sum_{d=1}^{7} \, \alpha_{d,y_d} \, \mathcal{L}_{ce}(\mathbf{p}_d, y_d),
\end{equation}
where $\mathbf{p}_d$ is the predicted score distribution for attribute dimension $d$, $y_d$ is the corresponding ground-truth label. In this way, the model is trained only on valid supervision while assigning larger learning emphasis to rare but informative score categories. 
Overall, our model training framework decouples supervision scale from supervision fidelity: it first learns broad attribute-level discrimination from inexpensive weak labels, and then improves reliability through consensus-based refinement under an imbalance-aware objective.

\subsection{RL Optimization with Adaptive Multi-Attribute Rewards}

DAT provides an effective reward signal for RL optimization in VTON. We use DAT as the reward model and optimize Qwen-Image-Edit~\cite{wu2025qwen} with Flow-GRPO~\cite{liu2026flow} as the policy optimization framework. The key problem is how to aggregate the seven attribute-level scores into a scalar reward for RL optimization. A fixed average or sum would treat all dimensions equally and may cause the policy to under-optimize persistent weak attributes.

To address this problem, we propose an adaptive multi-dimensional reward aggregation mechanism. We illustrate the training pipeline in Fig.~\ref{fig:method_rl}. Let $\mathbf{s}^{(n)}=[s_1^{(n)},\dots,s_7^{(n)}]$ denote the reward model scores for the $n$-th sampled result, and let $\mathcal{V}^{(n)}$ be the set of valid dimensions. We define the scalar reward as
\begin{equation}
R^{(n)} = \frac{\sum_{d \in \mathcal{V}^{(n)}} w_d^{(t)} \, s_d^{(n)}}{\sum_{d \in \mathcal{V}^{(n)}} w_d^{(t)}},
\end{equation}
where $w_d^{(t)}$ is the aggregation weight for dimension $d$ at iteration $t$.

We maintain an EMA of batch-wise scores for each dimension:
\begin{equation}
\bar{s}_d^{(t)} =
\begin{cases}
\mu_d^{(t)}, & t=1,\\
\lambda \bar{s}_d^{(t-1)} + (1-\lambda)\mu_d^{(t)}, & t>1,
\end{cases}
\end{equation}
where $\mu_d^{(t)}$ is the mean valid score of dimension $d$ in the current batch. Based on these running scores, we compute the short-board gap
\begin{equation}
g_d^{(t)} = \max_j \bar{s}_j^{(t)} - \bar{s}_d^{(t)},
\end{equation}
and define the adaptive weight as
\begin{equation}
a_d^{(t)} = \min\left(\tau,\, 1 + \gamma g_d^{(t)}\right).
\end{equation}
To optionally encode prior importance, we further introduce a prior weight $\pi_d$ and use
\begin{equation}
w_d^{(t)} = \pi_d \, a_d^{(t)}.
\end{equation}
This mechanism automatically upweights underperforming dimensions and encourages the policy to focus on current weaknesses.

Given scalar rewards $\{R^{(n)}\}_{n=1}^{G}$ for a sampled group of size $G$, we compute group-normalized advantages as
\begin{equation}
A^{(n)} = \frac{R^{(n)} - \mu_R}{\sigma_R + \epsilon},
\end{equation}
where $\mu_R$ and $\sigma_R$ are the mean and standard deviation of rewards within the group. The subsequent policy optimization follows the standard Flow-GRPO objective.

\section{Experiments}

\begin{table*}[t]
\centering
\small
\begin{tabular}{lccccccccc}
\toprule
& \multicolumn{3}{c}{In-domain} & \multicolumn{3}{c}{Out-of-domain} & \multicolumn{3}{c}{Overall} \\
\cmidrule(lr){2-4} \cmidrule(lr){5-7} \cmidrule(lr){8-10}
Method & Bal. Acc. $\uparrow$ & SROCC $\uparrow$ & PLCC $\uparrow$
       & Bal. Acc. $\uparrow$ & SROCC $\uparrow$ & PLCC $\uparrow$
       & Bal. Acc. $\uparrow$ & SROCC $\uparrow$ & PLCC $\uparrow$ \\
\midrule
Qwen3-vl-8b & 57.46\% & 0.2209 & 0.3506 & 58.86\% & 0.3461 & 0.3769 & 58.16\% & 0.2835 & 0.3637 \\
Qwen3.7-plus   & 69.76\% & 0.5186 & 0.5758 & 69.10\% & 0.3916 & 0.5299 & 69.43\% & 0.4551 & 0.5528 \\
GPT-5.5        & 74.97\% & 0.5258 & 0.5457 & 73.47\% & 0.4117 & 0.5440 & 74.22\% & 0.4688 & 0.5448 \\
Gemini-3.1 & \textbf{76.35\%} & 0.5744 & 0.5830 & 75.36\% & 0.4444 & 0.5595 & 75.85\% & 0.5094 & 0.5713 \\
\midrule
Ours           & 76.07\% & \textbf{0.5884} & \textbf{0.6174} & \textbf{77.03\%} & \textbf{0.5048} & \textbf{0.6126} & \textbf{76.55\%} & \textbf{0.5466} & \textbf{0.6150} \\
\bottomrule
\end{tabular}
\vspace{-0.1cm}
\caption{Qualitative comparison of different assessment models. We report balanced accuracy, SROCC, and PLCC on both in-domain sources and unseen out-of-domain sources. The best results are highlighted in \textbf{bold}.}
\label{tab:main_results}
\end{table*}

\begin{figure*}[!htbp]
\centering
\includegraphics[width=0.93\textwidth]{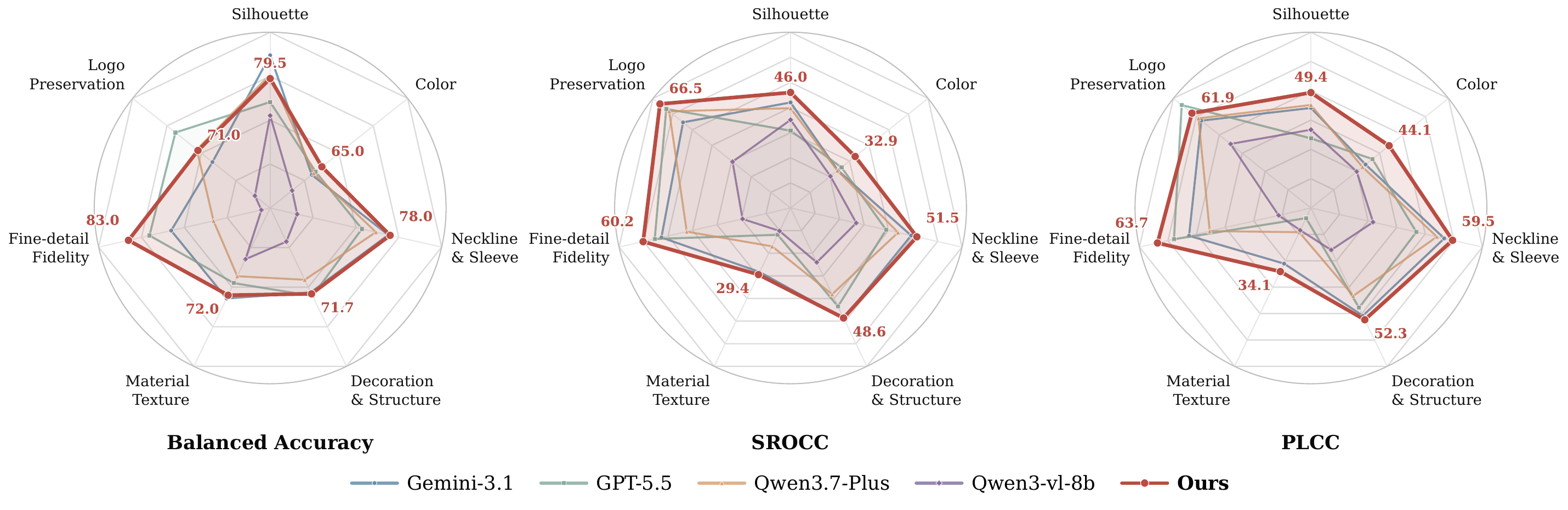} 
\vspace{-0.5cm}
\caption{Dimension-wise comparison of different assessment models in balanced accuracy, SROCC, and PLCC. Our method achieves a more balanced and consistent performance across fine-grained garment-consistency dimensions.}
\label{fig:radar}
\vspace{-0.5cm}
\end{figure*}

\subsection{Experimental Setup}

\subsubsection{Implementation Details.}

We initialize DAT from the pretrained Qwen3-VL-8B and fine-tune it with LoRA~\cite{hu2022lora} under our two-stage supervision strategy. We train the model for up to 16 epochs using AdamW~\cite{loshchilov2017decoupled} with a batch size of 8 and a learning rate of $1\times10^{-4}$. 
%
For reward-guided optimization, we choose the 4-step distilled Qwen-Image-Edit~\cite{qwen_image_edit_lightning} as the base model for fast training and leverage DAT to provide the reward signal for policy optimization via Flow-GRPO~\cite{liu2026flow}. We perform LoRA-based policy optimization with a learning rate of $1\times10^{-4}$ for five epochs. All experiments are conducted on 16 NVIDIA H20 GPUs.

\subsubsection{Datasets and Evaluation for Assessment Model.}

For DAT training, we collect 50K virtual try-on samples from an e-commerce platform, where each sample consists of a garment image, a person image, and a try-on result. The try-on results are generated by three virtual try-on systems: Nano Banana Pro~\cite{nanobanana_pro}, Qwen-Image-Edit-2511~\cite{wu2025qwen}, and an internally developed try-on model.

%
Following our proposed annotation protocol, we leverage Gemini-3.1~\cite{gemini_31_pro} to obtain seven dimension-wise garment consistency scores for each sample, providing large-scale supervision signals for assessment model learning. We further curate 10K high-quality annotations through multi-model voting among Gemini-3.1~\cite{gemini_31_pro}, Qwen3.7-plus~\cite{qwen37_plus}, and GPT-5.5~\cite{gpt55}.

To evaluate assessment models, we collect real-world garment and person images from an e-commerce platform and construct a test set with 1.6K samples. The try-on images are generated by six systems: Nano Banana Pro~\cite{nanobanana_pro}, Qwen-Image-Edit-2511~\cite{wu2025qwen}, the internally developed virtual try-on model, FireRed~\cite{firered2026rededit}, Nano Banana 2~\cite{nanobanana2} and GPT-Image2~\cite{gpt_image2}.
The first three sources overlap with the training data, while the remaining three are completely held out during training and are used to evaluate cross-source generalization. To obtain reliable ground-truth annotations, each test sample is independently evaluated by five expert annotators along the same seven garment-consistency dimensions, and the final labels are determined by majority voting.

For evaluation metrics, we use balanced accuracy~\cite{brodersen2010balanced}, where scores of 1--3 are treated as negative and scores of 4--5 as positive. To account for class imbalance, we average the accuracies on positive and negative samples:
\begin{equation}
\mathrm{BA} = \frac{1}{2} \left( \mathrm{Acc}_{\mathrm{pos}} + \mathrm{Acc}_{\mathrm{neg}} \right).
\end{equation}
We additionally report SROCC and PLCC to measure correlation with human annotations. For our seven-dimensional predictions, both metrics are computed separately for each dimension and then averaged across dimensions.

\subsubsection{Datasets and Evaluation for Reward-guided Optimization.}
For reward-guided optimization, we collect 10,000 training samples from an e-commerce platform and introduce Tryon-RLEval, a benchmark with 300 randomly sampled, disjoint images for evaluation.
For evaluation metrics, we report our proposed seven-dimensional scores to show the effectiveness of reward-guided optimization. 
Furthermore, we include the commonly used FID metric to measure distribution-level realism between person images and generated images, and conduct blind user studies to provide a reliable and primary evaluation of try-on quality.
We further evaluate our method on the out-of-domain StreetTryon~\cite{cui2025street} dataset and follow its evaluation protocol by employing the FID metric under four different try-on settings.

\begin{table*}[!t]
\centering
\setlength{\tabcolsep}{4.5pt} 
\begin{tabular}{lccccccc>{\centering\arraybackslash}p{1.7cm}}
\toprule
Method
& \makecell{Silhouette $\uparrow$} 
& \makecell{\hspace{4pt}Color $\uparrow$\hspace{4pt}}
& \makecell{Neckline / \\ Sleeve $\uparrow$}
& \makecell{Decoration / \\ Structure $\uparrow$} 
& \makecell{Material \\ Texture $\uparrow$} 
& \makecell{Fine-Detail \\ Fidelity $\uparrow$}
& \makecell{\hspace{4pt}Logo \\ Preserv. $\uparrow$\hspace{4pt}} 
& \makecell{\hspace{4pt}FID $\downarrow$\hspace{4pt}} \\
\midrule

Qwen-Image-Edit & {4.04} & {3.68} & {3.63} & {3.08} & {3.61} & {3.05} & {2.23} & {92.27} \\

+ RL Optimization
& \textbf{4.24\makebox[0pt][l]{\hspace{2pt}\textcolor{red}{\scriptsize(+0.20)}}}
& \textbf{4.01\makebox[0pt][l]{\hspace{2pt}\textcolor{red}{\scriptsize(+0.33)}}}
& \textbf{3.85\makebox[0pt][l]{\hspace{2pt}\textcolor{red}{\scriptsize(+0.22)}}}
& \textbf{3.44\makebox[0pt][l]{\hspace{2pt}\textcolor{red}{\scriptsize(+0.36)}}}
& \textbf{3.88\makebox[0pt][l]{\hspace{2pt}\textcolor{red}{\scriptsize(+0.27)}}}
& \textbf{3.52\makebox[0pt][l]{\hspace{2pt}\textcolor{red}{\scriptsize(+0.47)}}}
& \textbf{2.80\makebox[0pt][l]{\hspace{2pt}\textcolor{red}{\scriptsize(+0.57)}}}
& \textbf{90.91\makebox[0pt][l]{\hspace{2pt}\textcolor{gaincolor}{\scriptsize(-1.36)}}}\\

\bottomrule
\end{tabular}
\vspace{-0.1cm}
\caption{Quantitative comparison on Tryon-RLEval across the seven garment-consistency scores and FID.}
\vspace{-0.1cm}
\label{tab:tryon_model}
\end{table*}

\begin{table*}[!t]
\centering
\renewcommand{\arraystretch}{1.0}
\setlength{\tabcolsep}{6pt}
\small
\begin{tabular}{lcccc}
\toprule
Method
& Shop-to-Street & Model-to-Model & Model-to-Street & Street-to-Street (Top) \\
\midrule
Qwen-Image-Edit & 46.77 & \textbf{11.30} & 42.65 & 38.84 \\
+ RL Optimization & \textbf{46.35} & 11.61 & \textbf{41.92} & \textbf{38.03} \\
\bottomrule
\end{tabular}
\vspace{-0.1cm}
\caption{ Comparison on the StreetTryOn~\cite{cui2025street}  across four settings: Shop-to-Street, Model-to-Model, Model-to-Street, and Street-to-Street (Top). We adopt FID as the evaluation metric, where lower values indicate better performance.}
\vspace{-0.5cm}
\label{tab:streettryon_rl_fid}
\end{table*}

\subsection{Main Results on Assessment Modeling}


\begin{figure}[t]
\centering
\includegraphics[width=0.99\linewidth]{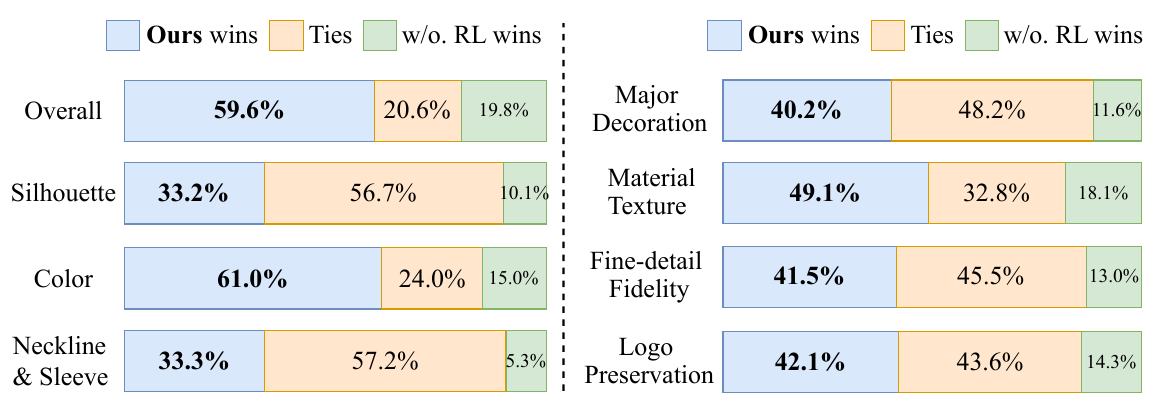} 
\vspace{-0.2cm}
\caption{Blind human study comparing the reward-guided model (\textit{Ours}) and the base model (\textit{w/o. RL}) across overall and seven-dimensional garment-fidelity scores.}
\label{fig:user_study}
\vspace{-0.5cm}
\end{figure}

We compare our assessment model with the baseline model Qwen3-VL-8B and several leading proprietary models, including Gemini-3.1~\cite{gemini_31_pro}, Qwen3.7-plus~\cite{qwen37_plus}, and GPT-5.5~\cite{gpt55}. 
Quantitative results on the evaluation test set are shown in Tab.~\ref{tab:main_results}. Overall, our method achieves the best performance among all competing approaches. Compared with the baseline Qwen3-VL-8B model, our model obtains nearly 20\% improvement while maintaining the same 8B-scale architecture. Moreover, it consistently outperforms proprietary models, demonstrating the effectiveness of our approach for fine-grained garment-consistency evaluation.
On in-domain sources, our method achieves the highest SROCC and PLCC while remaining competitive in balanced accuracy. More importantly, on out-of-domain sources, our method performs best across all metrics, indicating that the learned assessment model captures robust and transferable garment-consistency signals.

We provide Fig.~\ref{fig:radar} to illustrate dimension-wise performance across balanced accuracy, SROCC, and PLCC. Compared with proprietary methods, our method achieves the most balanced performance across all dimensions. While competing models excel on individual attributes, ours maintains consistent performance across fine-grained dimensions. This demonstrates that our multi-dimensional formulation captures garment consistency more comprehensively than global visual similarity.

\begin{figure*}[!htbp]
\centering
\includegraphics[width=0.95\textwidth]{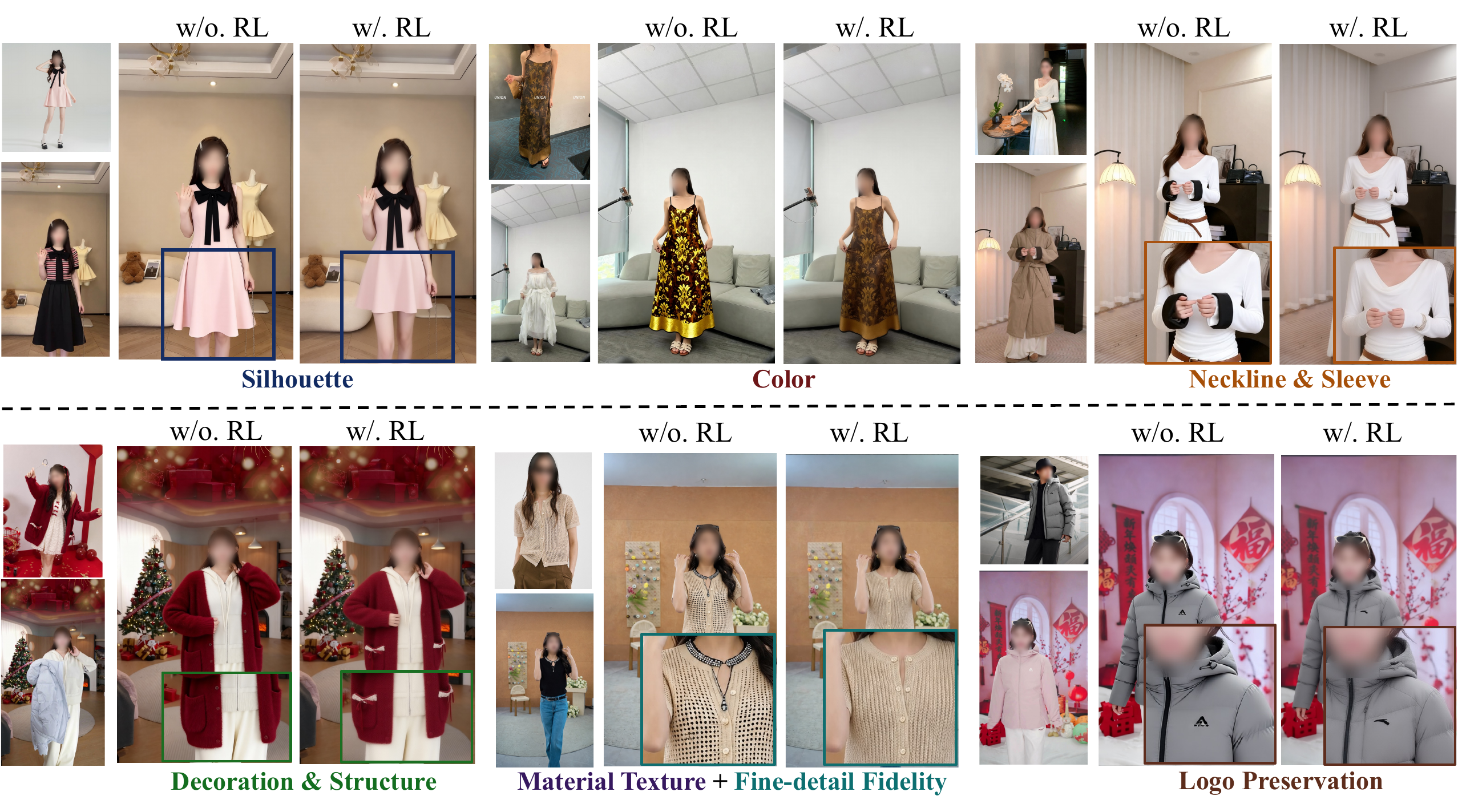} 
\vspace{-0.3cm}
\caption{Qualitative comparison of virtual try-on results without and with reward-guided RL optimization. RL optimization improves garment fidelity, leading to results that better match the reference garment in both global structure and local details.}
\label{fig:rl_image}
\end{figure*}

\begin{figure}[t]
\centering
\includegraphics[width=0.99\linewidth]{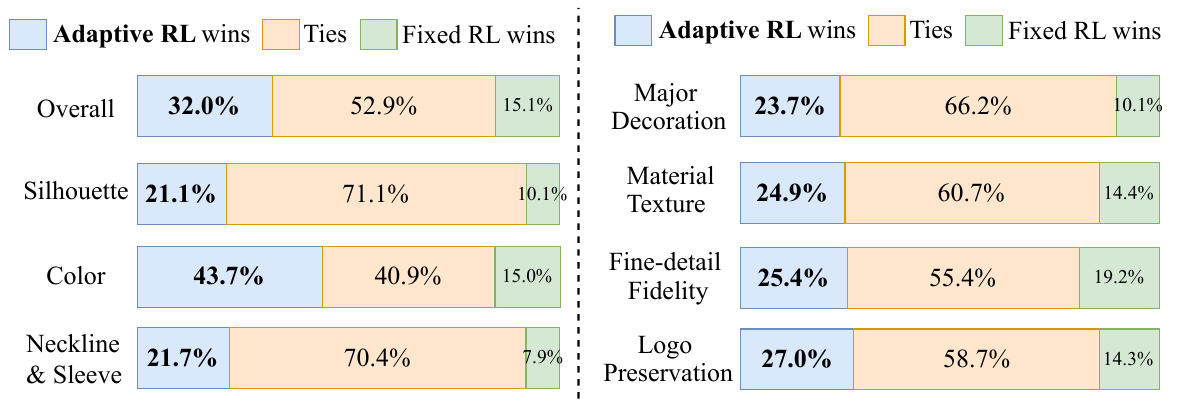} 
\vspace{-0.2cm}
\caption{Blind human study comparing adaptive-weighted RL (\textit{Adaptive RL}) and fixed-weighted RL (\textit{Fixed RL}) across overall and seven garment-fidelity dimensions.}
\label{fig:user_study_ablation}
\vspace{-0.5cm}
\end{figure}

\begin{table*}[t]
\centering
\setlength{\tabcolsep}{7pt} 
\begin{tabular}{lccccccc>{\centering\arraybackslash}p{1.6cm}}
\toprule
Method
& \makecell{Silhouette $\uparrow$} 
& \makecell{\hspace{4pt}Color $\uparrow$\hspace{4pt}}
& \makecell{Neckline / \\ Sleeve $\uparrow$}
& \makecell{Decoration / \\ Structure $\uparrow$} 
& \makecell{Material \\ Texture $\uparrow$} 
& \makecell{Fine-Detail \\ Fidelity $\uparrow$}
& \makecell{\hspace{4pt}Logo \\ Preserv. $\uparrow$\hspace{4pt}} 
& \makecell{\hspace{4pt}FID $\downarrow$\hspace{4pt}} \\
\midrule

Fixed & {4.15} & {3.87} & {3.75} & {3.26} & {3.78} & {3.30} & {2.55} & {91.92} \\

Adaptive
& \textbf{4.24\makebox[0pt][l]{\hspace{2pt}\textcolor{red}{\scriptsize(+0.09)}}}
& \textbf{4.01\makebox[0pt][l]{\hspace{2pt}\textcolor{red}{\scriptsize(+0.14)}}}
& \textbf{3.85\makebox[0pt][l]{\hspace{2pt}\textcolor{red}{\scriptsize(+0.10)}}}
& \textbf{3.44\makebox[0pt][l]{\hspace{2pt}\textcolor{red}{\scriptsize(+0.18)}}}
& \textbf{3.88\makebox[0pt][l]{\hspace{2pt}\textcolor{red}{\scriptsize(+0.10)}}}
& \textbf{3.52\makebox[0pt][l]{\hspace{2pt}\textcolor{red}{\scriptsize(+0.22)}}}
& \textbf{2.80\makebox[0pt][l]{\hspace{2pt}\textcolor{red}{\scriptsize(+0.25)}}}
& \textbf{90.91\makebox[0pt][l]{\hspace{2pt}\textcolor{gaincolor}{\scriptsize(-1.01)}} }
\\

\bottomrule
\end{tabular}
\vspace{-0.1cm}
\caption{Ablation study on different reward weighting strategies on Tryon-RLEval.}
\label{tab:rl_ablation}
\vspace{-0.3cm}
\end{table*}

\subsection{Main Results on Reward-Guided Optimization}


We further evaluate whether DAT can serve as an effective reward signal for RL optimization. Specifically, we compare the base Qwen-Image-Edit model with its reward-guided RL-optimized model from three perspectives:
(1) As shown in Tab.~\ref{tab:tryon_model}, we report our proposed seven-dimensional garment-consistency scores on Tryon-RLEval, yielding an average gain of 0.35, to show the effectiveness of reward-guided training. Meanwhile, the improved FID score indicates improved alignment with the learned garment-fidelity without sacrificing overall image quality.
(2) To further evaluate the improvement in try-on quality brought by reward-guided optimization, we conduct a blind human study on Tryon-RLEval. We invite 10 expert annotators for the user study. For each sample, annotators are given the garment image and two try-on results generated by the base model and the reward-guided model, respectively, and are asked to judge which result is closer to the reference garment from eight aspects: overall similarity, silhouette, color, neckline and sleeve shape, decoration and structure, material texture, fine-detail fidelity, and logo preservation. As shown in Fig.~\ref{fig:user_study}, reward-guided optimization improves performance consistently across all dimensions. Such experimental results confirm that the learned reward effectively promotes garment-faithful generation.
(3) Additionally, as shown in Tab.~\ref{tab:streettryon_rl_fid}, on the public StreetTryOn~\cite{cui2025street} dataset, reward-guided optimization improves FID over the base model.

\paragraph{Qualitative analysis.}
Fig.~\ref{fig:rl_image} shows representative examples before and after reward-guided optimization. The optimized model better preserves garment characteristics across our proposed seven dimensions, which is consistent with both the human study and quantitative results.

\subsection{Ablation Study}




\paragraph{Ablation on assessment model design.}
We conduct an incremental ablation study to demonstrate the reasonability of the training paradigm of DAT, with results presented in Tab.~\ref{tab:ablation_reward}. Compared with the baseline Qwen3-VL-8B, our Stage 1 model trained with coarse labels already achieves substantial improvements. Adding weighted CE consistently improves performance, highlighting the importance of imbalance-aware optimization for learning fine-grained garment consistency.
Introducing Stage 2 brings additional gains, as higher-quality labels generated by multi-model voting reduce annotation noise and provide more reliable supervision than coarse labels alone. 



\begin{table}[t]
\centering
\small
\begin{tabular}{cccccc}
\toprule
S1 & WCE & S2 & Bal. Acc. $\uparrow$ & SROCC $\uparrow$ & PLCC $\uparrow$ \\
\midrule
& & & 58.16\% & 0.2835 & 0.3637 \\
\checkmark &  &  & 74.28\% & 0.4872 & 0.5520 \\
\checkmark & \checkmark &  & 75.62\% & 0.5185 & 0.5632 \\
\checkmark & \checkmark & \checkmark & \textbf{76.07\%} & \textbf{0.5884} & \textbf{0.6174} \\
\bottomrule
\end{tabular}
\vspace{-0.1cm}
\caption{Ablation study on assessment model design. Stage 1 (S1) uses coarse labels. Weighted CE (WCE) assigns higher weights to underrepresented dimensions and hard samples. Stage 2 (S2) refines the model with higher-quality labels from multi-model voting.}
\vspace{-0.5cm}
\label{tab:ablation_reward}
\end{table}

\paragraph{Ablation on RL design.}

We conduct detailed experiments to investigate the effect of different reward weighting strategies in reward-guided optimization, including adaptive weighting and fixed weighting (average across all dimensions).
First, we report dimension-wise reward scores in Tab.~\ref{tab:rl_ablation}. Compared with fixed weighting, adaptive weighting achieves higher reward scores across all seven garment-consistency dimensions while maintaining better FID performance.
We further present a blind user study comparing adaptive and fixed weighting in Fig.~\ref{fig:user_study_ablation}. The results show that adaptive weighting consistently improves try-on quality, validating its effectiveness.
We attribute this to the fact that fixed weighting tends to bias optimization toward easier dimensions, whereas our adaptive strategy emphasizes lagging dimensions based on current performance, leading to more balanced garment-fidelity improvements.



\section{Conclusion}

In this paper, we propose a fine-grained assessment framework for garment consistency in virtual try-on. Our method decomposes garment fidelity into seven interpretable dimensions, enabling more informative attribute-level evaluation than conventional assessment metrics. Experiments on a manually annotated test set show that our model outperforms strong proprietary judges, including Gemini-3.1, Qwen3.7-plus, and GPT-5.5, in both binary accuracy and correlation with human annotations. Moreover, when used as a reward signal for reinforcement learning optimization of Qwen-Image-Edit, our model improves garment preservation and produces more faithful try-on results. These findings highlight the value of domain-specialized, interpretable, and fine-grained assessment for both evaluating and improving virtual try-on systems.

\bibliography{aaai2027}

\clearpage
\includepdfset{pagecommand={\thispagestyle{empty}}} 


\section*{Supplementary material}

# Supplementary Material: Beyond Global Realism: Virtual Try-On Evaluation and Optimization with Dimension-wise Garment Fidelity Assessment

**Kaidong Zhang[1]*, Yukang Ding[1]*, Xiaoyu Liu[2], Ying Chen[1]†**

[1]Taobao & Tmall Group of Alibaba, [2] Harbin Institute of Technology
zhangkaidong.zkd@taobao.com

## Details of the Dimension-wise Garment Assessment Protocol of DAT

To obtain a more diagnostic and interpretable evaluation of virtual try-on fidelity, DAT (**D**imension-wise **A**ssessment framework for virtual **T**ry-on) compares each source garment image with its corresponding generated virtual try-on image along seven dimensions: *silhouette*, *color*, *neckline and sleeve shape*, *major decoration and structure*, *material texture*, *fine-detail fidelity*, and *logo preservation*. DAT is designed to assess **garment-level consistency**, rather than overall visual plausibility alone, since a generated result may appear realistic while still failing to preserve product-defining attributes of the original garment. In this section, we present the details of these dimensions.

The seven dimensions capture complementary aspects of garment fidelity. *Silhouette* and *color* measure preservation of the garment's core identity. *Neckline and sleeve shape* and *major decoration and structure* evaluate key design elements that define garment style. *Material texture* assesses whether the perceived fabric characteristics are faithfully rendered. *Fine-detail fidelity* measures the preservation of small but semantically meaningful non-logo elements, while *logo preservation* is treated separately because logos are particularly sensitive to distortion and are of practical importance in commercial settings.

Each dimension is rated on a **five-point scale (1–5)**, where higher scores indicate better consistency between the source garment and the generated result. In general, **5** denotes near-perfect preservation, **4** indicates minor deviation, **3** corresponds to partial consistency with noticeable discrepancies, **2** reflects severe inconsistency, and **1** denotes complete failure. Dimensions that are not applicable to a given sample are marked as **N/A** and excluded from dimension-specific analysis. In particular, *fine-detail fidelity* is evaluated only when the garment contains relevant non-logo fine-grained elements (e.g., buttons, zipper pulls, embroidery, or small labels), and *logo preservation* is evaluated only when the source garment contains a clearly visible logo on the garment itself.

**Silhouette.** Evaluates whether the overall cut, length, proportion, and garment category are preserved. A score of **5** indicates exact preservation of the garment silhouette; **4** indicates slight proportional deviation that does not affect recognition; **3** indicates noticeable silhouette change while remaining within the same broad garment category; **2** indicates a clear change in garment type or core silhouette; and **1** indicates complete failure or severe structural corruption.

**Color.** Evaluates consistency of the garment's main color appearance. A score of **5** indicates an almost exact match in hue, saturation, and brightness; **4** indicates minor deviation likely caused by illumination or rendering effects; **3** indicates obvious color difference but within a similar color family; **2** indicates incorrect major color or hallucinated patterns; and **1** indicates chaotic or corrupted color rendering.

**Neckline and sleeve shape.** Evaluates whether the neckline type and sleeve type are preserved. A score of **5** indicates correct preservation of both; **4** indicates minor deviations in shape or proportion; **3** indicates partial resemblance with noticeable confusion in local design details; **2** indicates an incorrect basic neckline or sleeve type; and **1** indicates severe generation artifacts in the corresponding region.

**Major decoration and structure.** Evaluates the preservation of important decorative and structural components, such as printed pattern, pockets, belts, visible zippers, lace, paneling, and closure structure. A score of **5** indicates that all key elements are correctly preserved; **4** indicates minor deviations in location, shape, or styling; **3** indicates noticeable replacement or simplification of major elements; **2** indicates missing, hallucinated, or functionally altered structural components; and **1** indicates severe corruption of the relevant region.

**Material texture.** Evaluates whether the perceived fabric category and texture expression are faithfully rendered. A score of **5** indicates highly consistent material appearance; **4** indicates the correct material category with slightly weakened texture expression; **3** indicates a visually similar but mismatched material impression; **2** indicates clearly incorrect material type; and **1** indicates no meaningful texture expression or strong artificial artifacts.

**Fine-detail fidelity.** Evaluates the preservation of non-logo fine-grained elements, such as buttons, zipper pulls, embroidery, and small labels. A score of **5** indicates accurate

*These authors contributed equally.
†Corresponding author.

preservation of number, shape, position, and appearance; **4** indicates minor distortion; **3** indicates noticeable inconsistency or distortion; **2** indicates missing or incorrect elements; and **1** indicates severe corruption of the fine-detail region. This dimension is marked as **N/A** when no relevant fine-grained elements are present.

**Logo preservation.** Evaluates whether a visible garment logo is preserved in terms of shape, text, color, proportion, and position. A score of **5** indicates faithful and legible reproduction; **4** indicates slight distortion but clear recognizability; **3** indicates approximate preservation with substantial blur or deformation; **2** indicates loss of brand identity despite approximate placement; and **1** indicates complete disappearance or hallucination of an incorrect logo. This dimension is marked as **N/A** when no visible logo is present on the garment.

Overall, this fine-grained protocol provides a more informative benchmark than a single holistic score, as it explicitly reveals which aspects of garment fidelity are preserved well and which remain challenging for current virtual try-on methods.

## Training Data Filtering

Our training data are collected from a large-scale e-commerce platform and consist of two image sources: **garment images** and **person images**. Our garment images are not flat-lay product images (i.e., a single piece of clothing shown on a white background); rather, they are in-the-wild images. In other words, our garment images include a large number of instances where models are wearing the target product. Person images are obtained from the live-streaming rooms of the e-commerce platform channels. To improve data quality and reduce noisy supervision, we apply a multi-stage filtering pipeline before constructing training pairs.

First, we remove corrupted or invalid samples, including damaged garment images and damaged person images that cannot be reliably decoded or displayed. Second, we discard low-quality samples whose image resolution is lower than 720p, as well as images that are excessively blurry. To identify blurred images, we employ a blur detection operator (Pech-Pacheco et al. 2000) and remove samples that fall below a predefined sharpness threshold.

Third, for garment images, we use a vision-language model (VLM) to identify and exclude samples that contain multiple distinct products in a single image. Such cases are common in e-commerce data, where one product image may simultaneously display multiple variants, such as different colors or different styles of the same item. Since our goal is to establish an unambiguous correspondence between the source garment and the target try-on result, these multi-product samples are removed from the training set.

Fourth, for person images, we use the face detector provided by MediaPipe, an open-source cross-platform ML framework developed by Google, to filter out images with multiple faces. Samples containing multiple persons are excluded, as they introduce ambiguity regarding which person should be associated with the target garment during training.

After filtering, we obtain a cleaner set of product images and model images suitable for synthetic pair construction.

## Synthetic Pair Construction

After the filtering stage, we randomly pair a garment image with a person image and use an image editing model to synthesize a virtual try-on result. We utilize Nano Banana Pro (Google 2025b), Qwen-Image-Edit-2511 (Wu et al. 2025) and an internally developed try-on model to generate the try-on results. The editing instruction for each paired sample is automatically generated by Gemini-3.1 (Google 2025a), which produces a single English prompt describing how the garment in the model image should be replaced by the garment shown in the product image.

The prompt generation is designed to enforce high-fidelity product transfer. In particular, the generated instruction explicitly specifies the object to be replaced in the model image, identifies the target garment in the product image, and emphasizes that the edited result should preserve the target garment's fidelity, while keeping the other parts of the person image unchanged.

To avoid ambiguity in complex scenes, the prompt generator follows several additional rules. If the person in either image is wearing multiple layers of clothing, only the outermost garment is replaced. If the person in the model image is holding a hanger with a garment, only the held garment is replaced. These constraints help reduce editing ambiguity and improve the reliability of the synthesized virtual try-on data. The full prompt template is shown below.

**ROLE & GOAL**
You are an expert AI assistant specializing in generating precise, unambiguous English instructions for an image editing model. Your goal is to create a prompt that guides the editing model to perform a product swap in an e-commerce live streaming image.

**TASK**
Given two images, generate a single, clear text instruction for the image editing model.

**INPUTS**
1. **Image 1:** An image of a model/host displaying a product (the source image).
2. **Image 2:** A standalone image of the new product to be swapped in (the reference image).

**OUTPUT FORMAT**
Your output must be a single line of text containing the English editing instruction.

**CRITICAL RULES**
1. Identify the object to be replaced in Image 1 and describe the new object from Image 2.
2. Explicitly state that the new product appearance should be based on Image 2 or the reference image.
3. Clearly instruct the model to replace or swap the items.
4. If the person in Image 1 or Image 2 is wearing multiple garments, replace only the outermost garment. If the person in Image 1 is holding a hanger with a garment, replace only the held garment and do not replace the garment worn by the person.
5. Emphasize that the material, color, texture, details, and

any logos of the new product from Image 2 must be accurately preserved.
6. Do not change the person's face, expression, or gesture.

## User Study Protocol

We conduct a human preference study to compare the quality of virtual try-on models with or without RL optimization based on DAT. The goal of this study is to evaluate, from a garment-fidelity perspective, which generated result is more consistent with the garment image.

In each comparison instance, annotators are presented with a triplet of images composed of: (1) the garment image on the left, (2) a virtual try-on result generated by Method A in the middle, and (3) a virtual try-on result generated by Method B on the right. The two try-on results are generated by two different models, and their left-right assignment is randomized to avoid positional bias. Both try-on images are synthesized by image editing models that integrate the garment from the garment image onto a person image.

Annotators are asked to evaluate the two generated results by comparing the garment from each of them against that from the garment image. For each comparison instance, annotators complete eight annotation tasks, each corresponding to a different evaluation dimension. The objective is to determine which method produces a try-on result whose garment is more faithful to the garment image.

To improve annotation reliability, annotators are instructed to inspect each triplet at least twice before making any decision. For each task, annotators must select exactly one label. The available labels are Method A, Method B, and Same, except for dimensions that may be absent in the garment image, in which case an additional N/A-type option is provided.

The eight annotation tasks are defined as follows:

**Overall consistency.** Which result is more consistent with the garment image overall?
Options: **Method A / Method B / Same**

**Silhouette** Which result better matches the silhouette or overall garment shape in the garment image?
Options: **Method A / Method B / Same**

**Color.** Which result better matches the garment color in the garment image?
Options: **Method A / Method B / Same**

**Neckline and sleeve shape.** Which result better matches the neckline and sleeve shape in the garment image?
Options: **Method A / Method B / Same**

**Major decoration and structure.** Which result better matches the major decorative and structural elements of the garment in the garment image?
Options: **Method A / Method B / Same**

**Material texture.** Which result better matches the perceived garment material in the garment image?
Options: **Method A / Method B / Same**

**Fine-detail fidelity.** If the garment contains fine-grained elements, which result better preserves those details with respect to the garment image?
Options: **Method A / Method B / Same / No fine-grained elements**

**Logo preservation.** If the garment contains a visible logo, which result better preserves the logo with respect to the garment image?
Options: **Method A / Method B / Same / No visible logo**

This user study protocol enables direct pairwise comparison between different virtual try-on systems and provides fine-grained human judgments on which method better preserves the identity-defining attributes of the source garment.

 

\end{document}